\def\year{2022}\relax
\documentclass[letterpaper]{article} 
\usepackage{aaai22}  
\usepackage{times}  
\usepackage{helvet}  
\usepackage{courier}  
\usepackage[hyphens]{url}  
\usepackage{graphicx} 
\usepackage{natbib}  
\usepackage{caption} 
\DeclareCaptionStyle{ruled}{labelfont=normalfont,labelsep=colon,strut=off} 
\usepackage{amsfonts}
\usepackage{amsmath}
\usepackage{enumerate}
\newtheorem{defn}{Definition}
\usepackage{caption}
\usepackage{mathrsfs}
\usepackage[linesnumbered,ruled,vlined]{algorithm2e}
\usepackage{graphicx}
\usepackage{csquotes}
\usepackage[T1]{fontenc}
\usepackage{multirow}
\usepackage{makecell, caption}
\usepackage{tabularx}
\usepackage{varwidth}

\newlength\mylen
\newcommand\myinput[1]{%
	\settowidth\mylen{\KwIn{}}%
	\setlength\hangindent{\mylen}%
	\hspace*{\mylen}#1\\}

\newcommand{\LineNumbered}{%
	\setboolean{algocf@linesnumbered}{true}%
	\renewcommand{\algocf@linesnumbered}{\everypar={\nl}}}%

\let\oldnl\nl
\newcommand{\nonl}{\renewcommand{\nl}{\let\nl\oldnl}}

\newcolumntype{L}{>{\raggedright\arraybackslash}X}

\graphicspath{{figures/}}

\usepackage[noend]{algpseudocode}
\usepackage{amsmath}
\usepackage{mathtools}
\usepackage{xparse}
\usepackage{booktabs}

\usepackage{mathrsfs}

\usepackage{newfloat}

\DeclareDocumentCommand{\outAlgo}{ m O{\gets} }{%
	{\rlap{$#1$} \hspace{0.7cm}\hphantom{text}$#2$}%
}

\DeclareDocumentCommand{\jingAlgo}{ m O{-} }{%
	{\rlap{$#1$} \hspace{0.7cm}\hphantom{text}$#2$}%
}
\usepackage{todonotes}

\title{LIMREF: Local Interpretable Model Agnostic Rule-based Explanations for Forecasting, with an Application to Electricity Smart Meter Data}

\author{
    Dilini Rajapaksha\textsuperscript{\rm 1},
    Christoph Bergmeir\textsuperscript{\rm 1}
}
\affiliations{
    \textsuperscript{\rm 1}Faculty of Information Technology\\

	Monash University, Clayton Campus\\
	Building 6, 29 Ancora Imparo Way\\
	Monash University VIC 3800\\
    dilini.rajapakshahewaranasinghage@monash.edu,christoph.bergmeir@monash.edu 
    
}

\usepackage{bibentry}
\usepackage{enumerate}
\usepackage[shortlabels]{enumitem}

\begin{document}

\maketitle

\begin{abstract}
Accurate electricity demand forecasts play a crucial role in sustainable power systems. To enable better decision-making especially for demand flexibility of the end user, it is necessary to provide not only accurate but also understandable and actionable forecasts.
To provide accurate forecasts Global Forecasting Models (GFM) trained across time series have shown superior results in many demand forecasting competitions and real-world applications recently, compared with univariate forecasting approaches. 
We aim to fill the gap between the accuracy and the interpretability in global forecasting approaches.  
In order to explain the global model forecasts, we propose Local Interpretable Model-agnostic Rule-based Explanations for Forecasting (LIMREF), a local explainer framework that produces k-optimal impact rules for a particular forecast, considering the global forecasting model as a black-box model, in a model-agnostic way. 
It provides different types of rules that explain the forecast of the global model and the counterfactual rules, which provide actionable insights for potential changes to obtain different outputs for given instances. We conduct experiments using a large-scale electricity demand dataset with exogenous features such as temperature and calendar effects. 
Here, we evaluate the quality of the explanations produced by the LIMREF framework in terms of both qualitative and quantitative aspects such as accuracy, fidelity, and comprehensibility and benchmark those against other local explainers.

\end{abstract}

\section{Introduction}
Predicting electricity consumption and providing reasoning behind the predictions are important building blocks in the wider socio-economic challenge of transforming energy consumers to prosumers, needed to transform the energy sector to a carbon-emmission-free industry.
Forecasting the electricity demand on a smart meter level for particular households has been studied widely, and the effects of exogenous factors such as the temperature, daily, weekly, and yearly seasonalities, and other calendar effects, are well established. 
However, accurate energy forecasts are not enough to change the consumers' behaviour towards greater demand flexibility. We need to give understandable and actionable insights to the end user to enable behaviour change.

In the forecasting literature there are plenty of works on accurate electricity demand forecasting~\citep{porteiro2020electricity,blum2013electricity,taylor2003short,bedi2018empirical,al2018short}. The traditional energy demand forecasting methods are state-of-the-art univariate forecasting approaches such as ARMA, Trend Seasonal (TBATS) models and ARIMA models. The main limitation of these approaches is that they do not consider the information across multiple series. This cross-learning ability is particularly relevant when the time series are sharing key patterns and common structures among them.  
Global Forecasting Models (GFMs)~\citep{januschowski2020criteria} that are trained across all available time series have recently dominated traditional forecasting approaches in many cases including the M4 competition~\citep{makridakis2018statistical,Smyl2018-ak, smyl2020hybrid}, the M5 competition~\citep{makridakis2020m5}, various Kaggle competitions~\citep{bojer2020kaggle}, and also real-world applications in companies like Amazon~\citep{salinas2020deepar}, Walmart~\citep{bandara2019sales}, and Uber~\citep{laptev2017time}.

However, there is very limited work yet to explain the forecasts of GFMs. More general, to explain model predictions in machine learning the concept of ``local interpretability'' has been proposed first in the work of~\citet{Ribeiro2016-sp} which produces feature importance based explanations for individual predictions rather than providing explanations globally across all instances. Subsequently, researchers proposed multiple approaches which provide explanations for the global model predictions based on feature-importance~\citep{lundberg2017unified}, saliency maps~\citep{shrikumar2017learning,baehrens2010explain}, rule-based explanations~\citep{ribeiro2018anchors,rajapaksha2020lormika} and other methods~\citep{samek2021explaining}. Out of these approaches only~\citep{rajapaksha2020lormika} produces actionable counterfactual explanations. In machine learning there are some other counterfactual explanations approaches, e.g., minimum distance counterfactuals~\citep{wachter2017counterfactual}, feasible and least-cost counterfactuals~\citep{ustun2019actionable}, causally feasible counterfactuals~\citep{mahajan2019preserving,karimi2020model} and counterfactual explanations through data manifold closeness and sparsity~\citep{verma2020counterfactual,poyiadzi2020face}. 
However, these approaches are not directly applicable to GFMs and do typically not produce actionable explanations.

In this work, our aim is to propose a general methodology to explain the forecasts of a GFM. Then, we focus our method to the use case of electricity customers, to enable them to make better data-informed decisions using the generated explanations. 
%
In particular, we propose a framework called \textbf{LIMREF} \textbf{L}ocal \textbf{I}nterpretable \textbf{M}odel-agnostic \textbf{R}ule-based \textbf{E}xplanations for \textbf{F}orecasting which produces different types of rules as guidance to support electricity consumption planning.
Our proposed method has been developed for the 
FUZZ-IEEE Competition on Explainable Energy Prediction \citep{Triguero2021FUZZ}.
The problem posed in the competition is to provide accurate and self-explaining electricity demand predictions for a set of customers for each month over the coming year (January to December).
Our solution to the competition used a GFM based on Expectile Regression (ER)~\citep{sobotka2012geoadditive} to generate accurate forecasts. LIMREF then provides rule-based local explanations, where, in line with the local interpretability literature, local means an explanation of the global model forecast for each series/customer separately. 
To the best of our knowledge, LIMREF is the first model-agnostic algorithm that provides local rule-based explanations in time series forecasting. Moreover, the provided rules explain the decision and also provide counterfactual rules which are often important as actionable insights to change the outcomes. 

We evaluate our LIMREF framework and achieve competitive results on the competition dataset which contains half-hourly energy readings for 3248 smart meters, between January 2017 to December 2017, along with temperature data.  
Furthermore, we conduct both qualitative and quantitative experiments to assess the quality of the explanations and the global model, and are able to achieve competitive results.
In summary, the contributions of this paper are as follows.
(i) We propose a novel approach based on impact rules to generate local rule-based model-agnostic explanations for time series forecasting (ii) We further tailor the approach to an electricity demand prediction use case with global time series forecasting models. (iii) Our approach produces different types of rules, such as affirmative and counterfactual rules, which lead to actionable explanations.


\section{Background and theoretical foundations}
\label{sec:background}

In this section, we discuss the relevant background from association rule mining research with respect to impact rules where the target of the dataset is a numeric variable~\citep{webb2001discovering}, as well as the background of local interpretability and how to translate it to forecasting. 

\subsection{Impact Rule Mining}
\label{sec:imprule}

Association rule mining is a rule-based machine learning approach that was first introduced by~\citet{agrawal1993mining} to detect interesting combinations of input and output variables in large databases. In this context, rules can be defined as follows.

\begin{defn} \label{Def:AssociationRules}
Rule: A rule in our context follows the form $r_c = p \rightarrow q$. The left-hand side (LHS) of a rule, $p$, is also called the antecedent of the rule and consists of boolean conditions on feature values. The right-hand side (RHS) of a rule, $q$, is called the consequent of the rule and is the target value of the decision variable. 
\end{defn}

In association rules, the consequent is typically a class label. However, in our context of time series forecasting, the target outcome is a numeric value. A straightforward way to use class association rules for such a regression task, would be to use a discretized target. An advanced algorithm in this line is proposed in the work of~\citet{srikant1996mining}. 
However, \citet{webb2001discovering} argues that the ``provision of statistics relating to the frequency of one sub-range of values of a target variable does not directly address the issue of the impact on the general distribution of values for that variable.''
Therefore, to mitigate the above drawbacks, \citet{webb2001discovering} proposed impact rules where the segmentation of data is on the basis of a numeric value (e.g., mean, sum, median) rather than a distribution distorted by discretization. 
Thus, in impact rules, the impact of the target can be measured based on the following statistics. 

%
%

\begin{itemize}
	\item \textit{Coverage}: Proportion of cases which satisfy the LHS.
	\item \textit{Mean}: The mean value of the target for the instances covered by the LHS.
	\item \textit{Sum}: The sum of the values of the target for the instances covered by the LHS.
	\item \textit{Impact}: $\mathrm{impact}= \mathrm{sum}-\mathrm{mean}*\text{absolute\_coverage}$ Here, absolute coverage is the number of cases that fulfill the conditions of the LHS of the rule. This indicates the magnitude of the impact of the conditions in the LHS of the rule on the target beyond the sum expected if the LHS was independent of the target.
\end{itemize}


The main objective of discovering interesting rules is to identify the group that contributes most or least to a particular outcome. For example,  considering the application of electricity demand forecasting, the conditions which explain to increase or decrease the monthly electricity demand are considered as the most interesting rules. 
\citet{webb2001discovering} argues that identifying the group with a higher \textit{mean} value of the target is not equal to identifying the group which contributes most or least to the total, as the group may be small. As a result, \citet{webb2001discovering} concludes that rules generated based on aggregated measures such as \textit{sum} and \textit{impact} measures add greater interest generally. 
In this work, we consider the impact of the target based on the mean, which calculates the mean value of the target for the instances covered by the conditions of the LHS of the rule. 

When it comes to mining the rules efficiently, in our work we use the OPUS\_IR algorithm developed by \citet{webb2001discovering}. OPUS\_IR is a statistically sound rule-mining algorithm which overcomes many of the shortcomings of other rule mining techniques by substantially pruning the search space in each pruning action~\citep{webb2001discovering}. 
OPUS search has been successfully used in both classification~\citep{webb1994recent} and association rule-mining~\citep{webb2000efficient} tasks. Furthermore, it has effective control over the spurious correlations~\citep{Webb2011-cu}.
Most importantly, the algorithm achieves a dramatic reduction of the computational time compared to other techniques while providing efficient and complete rules.  


\subsection{Local Interpretability in Forecasting}
The concept of local interpretability in machine learning can be formally defined, and transfered to time series forecasting to explain GFM forecasts, as follows. 

Let $X \in \mathbb{R}^{n\times m}$ be a training dataset with $n$ instances and $m$ features, and let $\boldsymbol{z}$ be an $n$ dimensional vector of corresponding actual target values, and let $g$ be a given global black-box background model that produces predictions $\hat{z}$ from $m$ dimensional input vectors $\boldsymbol{x}$, that is, $g(\boldsymbol{x})= \hat{z}$, such that the error between $z$ and $\hat{z}$ is minimised. The goal of a local interpretability algorithm is to find an interpretable explainer model $e$, for example, a linear model, such that for a given instance $\boldsymbol{x}$ where the global model predictions $\hat{z}$ need to be explained, $e(\boldsymbol{x}) \approx \hat{z}$.
To develop $e$, we first define a neighbourhood for the instance to be explained $\boldsymbol{x}$ following two steps. First, we select the most similar instances to $\boldsymbol{x}$ using a distance function. After, we create a set of newly generated instances that lie in the neighbourhood. We call this new set $\tilde{X}$. We then apply $g$ to all these instances in $\tilde{X}$ to obtain the global model predictions $\tilde{z}$. Here, we obtain $\tilde{z}$ to identify the behaviour of the global model when the instances are similar to the instance to be explained. Thus, $e$ is trained on $\tilde{X}$ to resemble $\tilde{z}$ as the target predictions $\hat{z}$. To generate $\tilde{X}$, usually in machine learning the interpretability algorithms first filter the most similar instances from the original training set $X$ as instances that lie in the neighbourhood of $\boldsymbol{x}$, into a set that we call $X_{\textit{filt}}$. Then, $X_{\textit{filt}}$ is used to generate $\tilde{X}$ by using sampling procedures and by adding noise to the features of $X_{\textit{filt}}$.

When transferring the concept of local interpretability to global model forecasting, the global forecasting model $f$ is considered as the background model that has been trained on a set of time series $Y=\{y_1,\ldots,y_p\}$, where $p$ is the total number of time series in the dataset.
Let the global model forecasts of a particular time series $y$ be $\hat{y}$, where $f(y) = \hat{y}$. The goal of this work is to develop a local explainer $e$ which is an interpretable model, such that $e(y)\approx \hat{y}$.
In our approach LIMREF to generate the neighbourhood $\tilde{X}$, we first filter the most similar (according to a distance measure to be chosen) time series from the original training set $X$ as instances that lie in the neighbourhood of $\boldsymbol{x}$. These series comprise $X_{\textit{filt}}$. Then, $X_{\textit{filt}}$ is used to generate $\tilde{X}$ by using bootstrapping procedures on $X_{\textit{filt}}$.
Finally, the local explainer $e$ which is an interpretable surrogate model is fitted on $\tilde{X}$ to provide local explanations for the forecasts $y$ of the time series to be explained.


\section{Proposed LIMREF Approach}

\begin{figure*}[htb]
	\centering
	\includegraphics[width=0.72\textwidth]{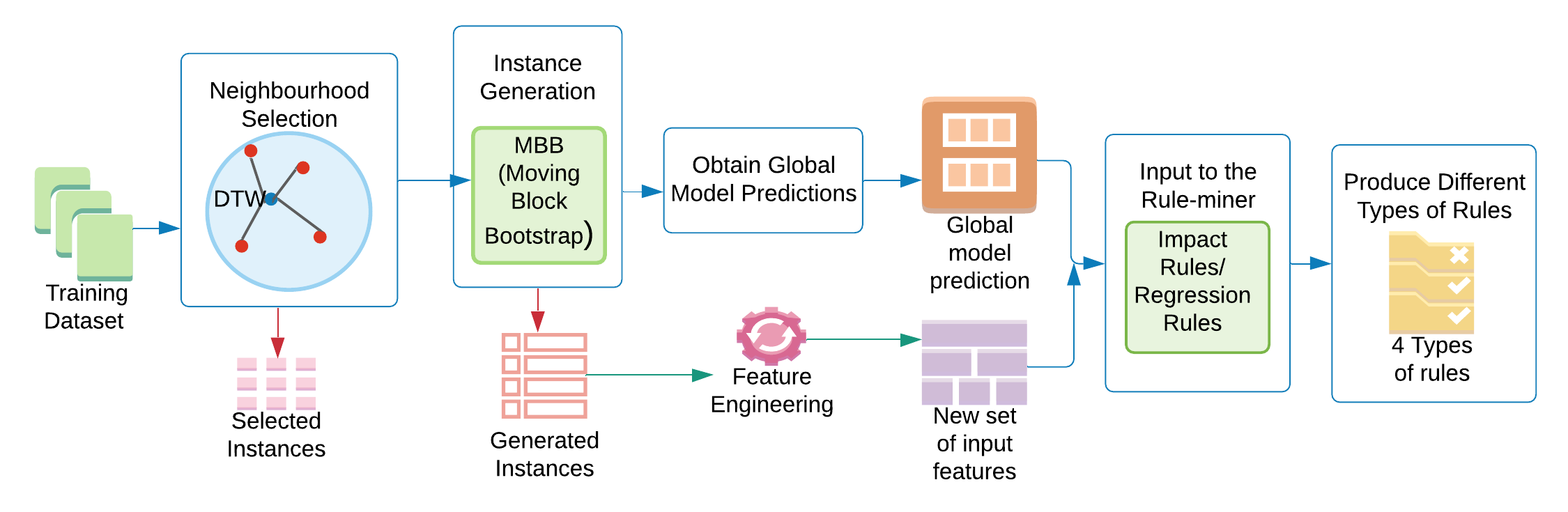}
	\caption{An overview diagram of our LIMREF architecture to generate different types of guidance in the form of rule-based explanations.}
	\label{fig:framework}
\end{figure*}

We first define a neighbourhood as time series that are similar to the series for which the forecasts are to be explained. Then, we use a bootstrapping technique to generate new time series from the ones in the neighbourhood. After, we generate global model forecasts for these series, namely for both the bootstrapped series and the original series in the neighbourhood. Finally, we perform impact rule mining based on the approach proposed in the work of~\citep{webb2001discovering} for the combined set of series and their global model forecasts to produce impact rules as the explanations. An overview diagram of the LIMREF architecture is shown in Figure~\ref{fig:framework}.
Our approach is furthermore summarized in Algorithm~\ref{Alg:limref} and the most important steps are discussed in the following.

\begin{algorithm}[tb]
	\caption{A Local Interpretable Model-agnostic Rule-based Explanations for Forecasting: LIMREF}\label{Alg:limref}
	\footnotesize
	\DontPrintSemicolon
	\SetKwInOut{Input}{Input}
	\SetKwInOut{Output}{Output}
	\Input{{\jingAlgo{Y_t}} {\text{training set without test data}}}
	\nonl\myinput{~~{\jingAlgo{x}} {\text{time series that need to be explained}}}
	\nonl\myinput{~~{\jingAlgo{f}} {\text{ a global demand forecasting model}}}
	\nonl\myinput{~~{\jingAlgo{N_\textit{synthetic}}} {\text{\# of new instances to be generated}}}
		\nonl\myinput{~~{\jingAlgo{N_\textit{filt}}} {\text{\# of nearest neighbours }}}
	\Output{$E_{\boldsymbol{x}} ~-$ Rule-based explanations for the forecast of the instance to be explained}
	\BlankLine
	\begin{algorithmic}[1]
		\SetAlgoLined
		\State{\outAlgo{X_{\textit{filt}}} $\text{SelectNearestNeighbours}(Y_{t},x,N_\textit{filt})$}
		\State{\outAlgo{X_{\textit{synthetic}}}$\text{GenerateFromNeighbours}(X_{\textit{filt}},N_\textit{synthetic})$}
		\State{\outAlgo{\tilde{X}}$ X_{\textit{filt}} \cup X_{\textit{synthetic}} $}
		\State{\outAlgo{f(\tilde{X})} $\text{GetPredictFromGlobalModel}(\tilde{X},f)$}
		\State{\outAlgo{\tilde{X}_\textit{proxy}} $\text{GetSurrogateFeatures}(\tilde{X})$}
		\State{\outAlgo{R_{\boldsymbol{x}}} $\text{DiscoverImpactRules}(\tilde{X}_\textit{proxy},f(\tilde{X}))$}
		\State{\outAlgo{E_{\boldsymbol{x}}} $\text{GenerateGuidance}(R_{{\boldsymbol{x}}}, {\boldsymbol{x}},f(\tilde{X}))$}
		\State \Return $E_{\boldsymbol{x}}$		
	\end{algorithmic}
\end{algorithm}

\subsubsection{Select the time series most similar to the time series to be explained}\label{subsec:selectInst}
This step corresponds to the $\text{SelectNearestNeighbours}(Y_{t},x,N_\textit{filt})$ function in Algorithm~\ref{Alg:limref}. Several distance measures have been proposed in the literature for time series. The most popular one is arguably Dynamic Time Warping (DTW)~\citep{keogh2005exact}, which allows a non-linear mapping of two vectors by minimizing the distance between them.
Following the work of~\citet{fu2008scaling} we perform mean normalization of the time series before applying DTW to achieve better results. 
After, we select the top $N_\textit{filt}$ instances with the smallest DTW distance to the time series in consideration. 
We use the implementation \texttt{dtw.distance\_matrix\_fast}~\citep{meert2020wannesm} in the \texttt{dtaidistance} package in Python. 

\subsubsection{Generate new series in the neighbourhood, and corresponding GFM predictions}

This corresponds to  $\text{GenerateFromNeighbours}(X_{\textit{filt}},N_\textit{synthetic})$ in Algorithm~\ref{Alg:limref}.
Following the local interpretability paradigm, we generate new synthetic instances in the neighbourhood. 
As our instances are time series, we employ a time series bootstrapping procedure, namely the procedure proposed by~\citet{bergmeir2016bagging}, Box-Cox and Loess-based decomposition bootstrap, to generate $N_\textit{filt}$ number of new series from each time series of the neighbourhood. We use the implementation in the \texttt{bld.mbb.bootstrap} function in the \texttt{forecast} package~\citep{hyndman2018forecast,hyndman2007automatic} in the R programming language.
We then generate global model forecasts for all series in the combined dataset $\tilde{X}$, which consists of the original series in the neighbourhood, $X_\textit{filt}$, and the bootstrapped series $X_\textit{synthetic}$, as target values for the local explainer.

\subsubsection{Generate surrogate feature set of inputs, and surrogate outputs (optional)}\label{subsub:surrogate_feature}

\begin{figure}[]
	\centering
	\includegraphics[width=\columnwidth]{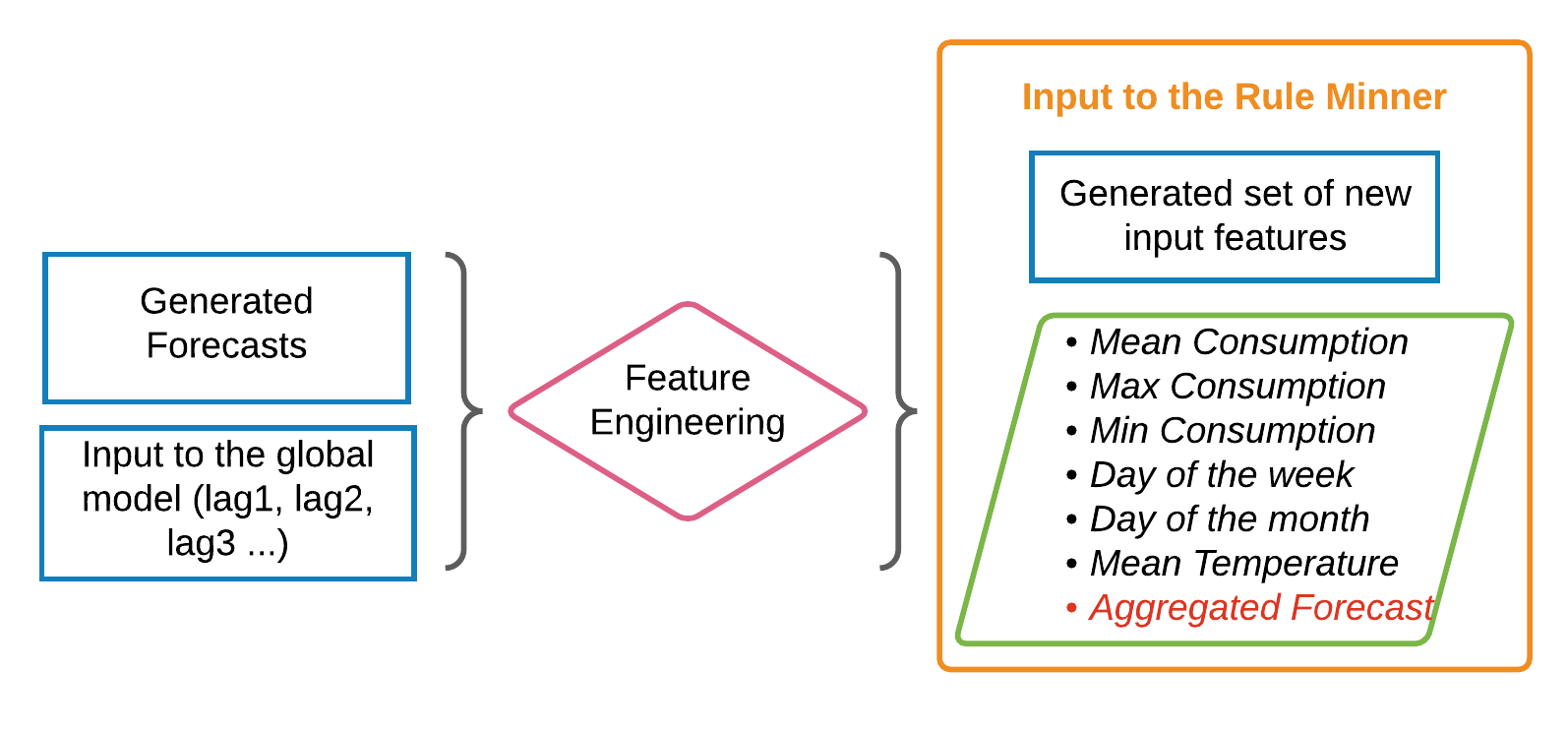}
	\caption{Feature engineering step for aggregated demand forecasting explanations.}
	\label{fig:featureEng}
\end{figure}

A drawback of local interpretability algorithms such as LIME~\cite{Ribeiro2016-sp} 
is that their explanations are feature importance values. Thus, if the original features are not interpretable, the explanations are not useful. In time series forecasting, we can be in a situation that the forecasts will be aggregated, e.g., the GFM provides daily forecasts that are subsequently aggregated to monthly forecasts. In this situation, explanations that use the daily input features and the daily forecasts are not useful.
%
%
Thus, once all neighbourhood series are selected and generated, and forecasts are produced, to provide meaningful explanations in such a situation when the forecasts will be later aggregated, we can perform customised feature engineering tasks, which is corresponding to the $\text{GetSurrogateFeatures}(\tilde{X})$ function in Algorithm~\ref{Alg:limref}. This effectively generates a surrogate task with both aggregated inputs and outputs, to yield relevant explanations. As an example, in our electricity demand forecasting case study, we aggregate the input electricity consumption data and the exogenous variables such as temperature to generate a new set of features such as mean electricity demand, minimum electricity demand, maximum electricity demand, mean temperature etc., also see Figure~\ref{fig:featureEng}.

\subsubsection{Generate impact rule-based explanations}
This step corresponds to the  $\text{DiscoverImpactRules}(\tilde{X}_\textit{proxy},f(\tilde{X}))$ function in Algorithm~\ref{Alg:limref}. We generate statistically sound impact rules using OPUS\_IR~\citep{webb2001discovering}.
%
As discussed before, by using OPUS\_IR, LIMREF is able to provide explanations that contribute large amounts to the total by using \textit{sum} and \textit{impact} measures as the filtering option. The filtering option based on the \textit{sum} will be useful if the target directly measures the end objective, such as the true profit from the electricity consumption. However, in the electricity demand study we aim to explain the monthly electricity consumption rate. Thus, we filter our rules based on the \textit{impact} since our target is an intermediate variable (electricity consumption). 

\subsubsection{Generate different types of guidance}

When explaining global classification models locally using association rules, \citet{rajapaksha2020lormika,rajapaksha2021sqaplanner} propose to use four types of explanations, defined by a contingency table of the LHS and RHS of the rule.   
Similarly, as the final step of our LIMREF framework, we categorize our rules into four types with respect to a contingency table of the LHS and RHS of the impact rules, which will finally produce different types of guidance.  This step corresponds to the function $\text{GenerateGuidance}(R_{{\boldsymbol{x}}}, {\boldsymbol{x}},f(\tilde{X}))$ in Algorithm~\ref{Alg:limref}.    
Then, we identify the rule with the highest impact for each type of guidance as the output of LIMREF.

The LHS of a rule is considered as \textit{true} if the antecedent of the rule applies to the surrogate feature values of the time series to be explained. In contrast to classification, in forecasting, the RHS of the rule and the global model forecast are not in categorical form. Consequently, we cannot verify directly if the consequent of the rule agrees with the forecast provided by the global model of the time series to be explained. 
Therefore, in our approach, the RHS of the rule is considered as \textit{true} if the global model forecast falls within one standard deviation of the neighbours whose feature values agree with the LHS of the rule.

Let us again consider an example from our monthly electricity demand forecasting scenario, where for the meter id $x$ the global model predicts the January electricity consumption would be ($p$). Assume from the neighbours of the time series to be explained, the mean of the global model forecasts whose features agree with the LHS of the rule is $\tilde{x}$, and $\delta$ is a constant value.
%
To explain this forecast LIMREF produces four types of rules which produces finally six types of guidance as follows. 


\textbf{Current Supporting rules ($\Re_{c\alpha}^+$ )}: The rules that support the forecast of the global model.
\emph{Definition}: IF LHS is true, THEN RHS is true ($p-\delta \leq \tilde{x} \leq p+\delta$).
\emph{Guidance type} \textbf{G1:} Explains the current practises that result in obtaining the predicted consumption $p$. 

\textbf{Current Contradicting Rules ($\Re_{c\beta}^-$ ):} The rules that contradict the prediction of the global model.
\\ \textbf{Type 1}
\emph{Definition}: IF LHS is true, THEN RHS is false, ($\tilde{x} > p+\delta$).
\emph{Guidance type} \textbf{G2:} Explains the current practises that indicate a result in increasing the predicted consumption $p$.
\emph{Benefits}: Observing the Current Contradicting Rules-Type 1, customers will be acknowledged about their ongoing risky practises which may result in an increase of the given prediction. Therefore, they should try to avoid those practises.
	\\ \textbf{Type 2}
\emph{Definition}: IF LHS is true, THEN RHS is false, ($\tilde{x} < p-\delta$).
\emph{Guidance type} \textbf{G3: }Explains the ongoing practises that result in decreasing the predicted consumption $p$.
\emph{Benefits}: Observing the Current Contradicting Rules-Type 2, the customers are informed about which factors they should not change.

\textbf{Hypothetically Supporting Rules ($\Re^{+}_{h\alpha}$) :} The rules would increase the probability of the prediction of the global model.
\emph{Definition}: IF LHS is false, THEN RHS is true ($p-\delta \leq \tilde{x} \leq p+\delta$).
\emph{Guidance type} \textbf{G4: } Explains the potential practices to follow to maintain the predicted consumption $p$.
	Determine the conditions that are currently not satisfied by $x$ (they are hypothetical) that would further result in obtaining the predicted consumption $p$. 
\emph{Benefits}: Observing the Hypothetically Supporting Rules, if the customers are happy about the predicted consumption $p$, following these conditions will further maintain $p$.

\textbf{Hypothetically Contradicting Rules (Counterfactual Rules, $\Re^{-}_{h\beta}$ )}: The rules that may invert the prediction of the global model.
	\\ \textbf{Type 1}
 \emph{Definition}: IF LHS is false, THEN RHS is false ($\tilde{x} > p+\delta$).
 \emph{Guidance type} \textbf{G5: }Explains the potential practises that result in increasing the predicted consumption $p$. Determine the conditions that are currently not satisfied by $x$ (they are hypothetical) that would further increase the predicted consumption $p$.
 \emph{Benefits}: Observing the Hypothetically Contradicting Rules-Type 1, the customers should try to avoid these risky practises which may increase the given prediction.
	\\ \textbf{Type 2}
\emph{Definition}: IF LHS is false, THEN RHS is false ($\tilde{x} < p-\delta$).
\emph{Guidance type} \textbf{G6: }Explains the potential practises that result in decreasing the predicted consumption $p$. Determine the conditions that are currently not satisfied by $x$ (they are hypothetical) that would decrease the predicted consumption $p$.
\emph{Benefits}: Observing the Hypothetically Contradicting Rules-Type 2, the customers should try to follow these non-risky practises which may decrease the given prediction.

\section{Application in Electricity Demand Forecasting}

In this section we describe the dataset, decision making approach and the evaluation criteria of the proposed LIMREF framework used for our experiments.

\subsection{Dataset}
We use the dataset provided in the FUZZ-IEEE competition on Explainable Energy Prediction 2021 \citep{Triguero2021FUZZ} for our experiments.  This dataset consists of historical half-hourly energy readings for 3248 smart meters for a one year period. The time series have different lengths, as customers may have joined at different times during the previous year. Moreover, together with the electricity consumption dataset, the competition provided access to the full-year temperature for each meter\_id or customer at a daily resolution, including the average temperature, minimum and maximum of each day.

\subsection{Electricity Demand Forecasting Model}
Our electricity demand forecasting and decision making approach consists of two major phases: (1) developing an accurate electricity demand forecasting model, and (2) generating local explanations using the LIMREF framework to explain the forecasts of the global electricity demand forecasting model. 


Though our paper presents an algorithm for local interpretability and accurate forecasting models are not the main interest for us, a forecasting model is needed to underpin our work. Our forecasting approach is based on the approach used by the team that won 4th place in the previous competition 
IEEE-CIS technical challenge on Energy Prediction from Smart Meter Data on the same dataset where the goal was to generate the most accurate forecasts, without regarding explainability. 
The following sections briefly describe the main components of the approach.


\subsubsection{Data pre-processing}

	\textit{Data Aggregation:} First we aggregate the half-hourly data into daily data to reduce the impact of the large proportion of missing values in the dataset. Therefore, our forecasting model will generate a daily forecast for 365 days ahead as the intended forecast horizon of this competition is 12 months. 
	\textit{Missing values imputation:} To do the further imputation of missing values, a seasonal imputation approach is performed by considering the median energy consumption of each day of the week of each household.
	\textit{Temperature estimation:} The competition dataset has provided temperatures only for the year 2017. A preliminary experiment which estimated temperature values with a Moving Block Bootstrapping procedure did not increase the accuracy of the global model significantly. Therefore, to estimate the temperature values for the forecasting period, we simply considered the same 2017 temperature values as 2018 values.  
	\textit{Normalizing data:} Since the electricity consumption varies across the different households as per the recommendations of~\citet{hewamalage2021recurrent} we performed a mean scaling normalization strategy. 

\subsubsection{Global forecasting model}
From preliminary experiments on the dataset, we found indications of systematic overestimation or underestimation of the final forecasts for yearly and monthyl predictions, respectively. To correct for this, we used Expectile Regression (ER)~\citep{sobotka2012geoadditive} as the GFM. Based on empirical findings we set a quantile of 0.57 for longer series and 0.39 for shorter series. To implement the ER global model we used the \texttt{expectreg}~\citep{expectreg2019} package in the R programming language.
%
We apply the moving window forecasting strategy with a window size of 20, and we iteratively generate daily forecasts for the whole next year, by feeding back the forecasts as inputs for the subsequent time steps. We finally aggregate into monthly and yearly forecasts.  


\subsection{Experiments}

In this section, we evaluate the LIMREF framework with two other interpretable models in both qualitative and quantitative aspects.

\subsubsection{Select interpretable models for baseline comparisons}
As LIMREF, to the best of our knowledge, is the first local rule-based model agnostic interpretability algorithm in forecasting, there are no baselines readily available. Therefore, as the baseline comparisons, we propose to obtain explanations and the forecasts by fitting existing interpretable algorithms such as linear regression (LR) and decision tree (DT) models.
The LR model produces coefficients that explain the most important features as the explanations for the global model forecasts whereas the DT produces a decision tree that we consider interpretable. The LR approach is comparable to LIME~\cite{Ribeiro2016-sp}, adapted to our time series application.

We fit all three local explainers (LIMREF, LR, and DT) on the same neighbourhoods. Therewith, as LIMREF, LR, and DT are deterministic algorithms, the results are directly comparable as the neighbourhood generation is then the only computation step that adds a certain degree of randomness to the explanations.

For the LR and DT models we use a cross-validation approach to select the best models, and we use the implementations from the \texttt{glmnet} package~\citep{glmnet2011} and the \texttt{rpart} package~\citep{rpart2019} in R, respectively.  

\subsubsection{Parameter setup of the local explainers}
Following computational feasibility considerations, we set $N\_{filt}$ to 50 which is the number of neighbours from the training set and we set $N_\textit{synthetic}$ to 100, which is the number of synthetic instances generated by the bootstrapping procedure from each selected neighbour of the time-series to be explained. Therewith, for 3428 households the size of the bootstrapped dataset is 3428 $\times$ 50 $\times$ 100, so that we consider over 17 million time series in total in the experiments. 

\subsection{Evaluating local explainer models}
Following the literature on local explainability \citep{yang2019bim,carvalho2019machine}, we evaluate our local explainers based on the comprehensibility, fidelity, and the accuracy. 
Based on a qualitative analysis we evaluate the comprehensibility of the explanations. Then, based on a quantitative analysis the fildelity and the accuracy of the local explainers are examined, which assesses how well the explainer approximates the forecasts of the GFM and the real values, respectively. 

\subsubsection{Qualitative analysis of the local explainers}
In this section, we illustrate with an example the comprehensibility of LIMREF, LR, and DT local explainers. We consider a household where the electricity demand forecast of the global model (ER) for the month of February is 129.27kWh. The feature values of the time series that needs to be explained are as follows.

\begin{table}[!tbh]
	\centering
	\caption{Original Values of the Instance to be Explained} 
	\label{tab:summary_datasets}
	\scalebox{0.8}{
		\begin{tabular}{cccccc}
			\hline
			Mean\_cons&Max\_cons& Min\_cons& Temp &Month & ER Forecast \\ 
			\hline
			16.22kWh &30.75kWh  & 0kWh	&6.01C& Feb&568.93kWh\\ 
			\hline
		\end{tabular}
	}
\end{table}


In the following, we discuss the possibilities of changes to the electricity consumption of the given instance for the month of February using the local explainer models. 

\paragraph{What are current practises that result in obtaining the predicted consumption 568.93kWh? }
This question can be answered using the current supporting rules $\Re_{c\alpha}^+$ of the LIMREF algorithm. 
\textit{$\Re^+_{{c\alpha\_}{{\text{LIMREF}}}}:$  Your predicted consumption is 568.94kWh. Because you have mean consumption > 14.74kWh.} 

\textit{Advantage}: This type of rule explains the current practices that result in obtaining the predicted consumption. Hence, having mean electricity consumption greater than 14.74kWh in the past 20 days is why the predicted consumption is 568.94kWh.


\begin{table}[!tbh]
	\centering
	\caption{Coefficients of local explainer LR model} 
	\label{tab:lr_exp}
	\scalebox{0.8}{
		\begin{tabular}{ccccc}
			\hline
			Mean\_cons & Max\_cons & Min\_cons & Temp & Month \\ 
			\hline
			28.4 &1.26  & -6.86 & 24.14& -0.52 \\ 
			\hline
		\end{tabular}
	}
\end{table}

To answer the same question, the LR explainer provides the feature importance in the form of the coefficients of the local linear model. Accordingly, Table~\ref{tab:lr_exp} shows that for the particular household the mean consumption and the temperature factors positively support the predicted consumption of 568.94kWh.   

\begin{figure}[htb]
	\centering
	\includegraphics[scale=0.2]{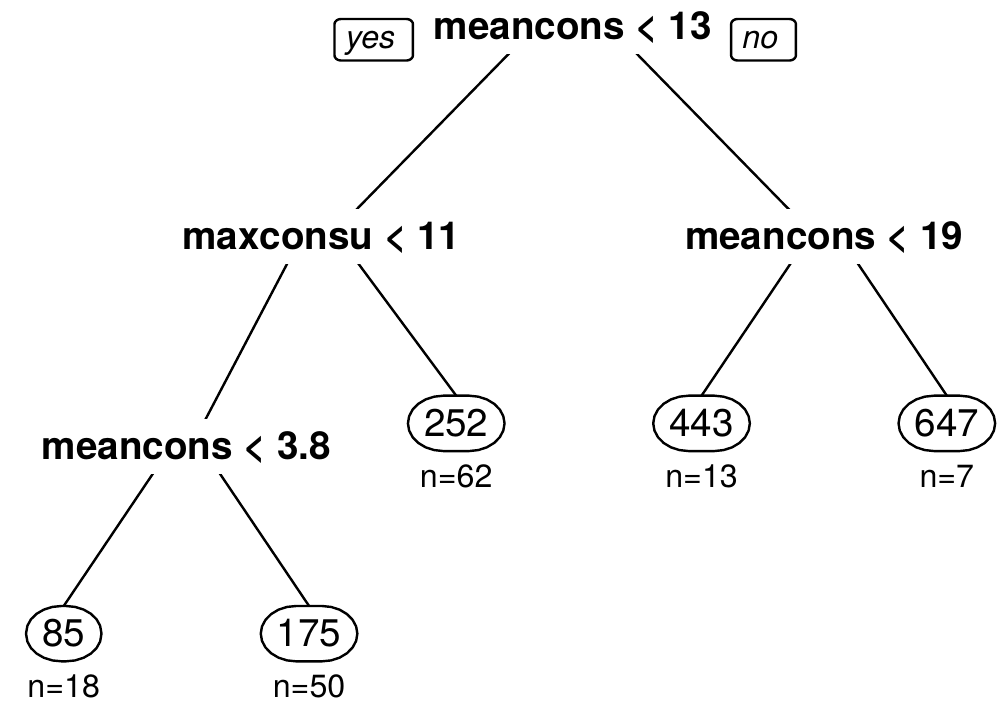}
	\caption{ Local interpretable explanation of DT model}
	\label{fig:dt_pic}
\end{figure}
The DT explainer provides a decision tree as the explanation. According to Figure~\ref{fig:dt_pic}, the DT explainer suggests that having mean consumption between 13kWh and 19kWh supports to have the predicted consumption of 568.94kW in February.


\paragraph{What are the current conditions which result in increasing or decreasing the predicted consumption 568.94kWh? }
To answer this question LIMREF is providing ($\Re_{c\beta}^-$), as follows:
\textit{The conditions that currently exist that indicate an opportunity of a decreased consumption by 170.45kWh for the particular month are 5.95 < average temperature $\leq$ 6.01.} 

\textit{Advantage}: Observing the Current Contradicting Rules-Type 2, the customers are informed about which factors they should not change. In particular, the customer understands that having an average temperature between 5.95C  and  6.01C, there can be a decreased consumption of 170.45kWh. In this case, the rules are not actionable since the outside temperature is not manageable.

To answer the same question with LR, Table~\ref{tab:lr_exp} shows that the minimum consumption and the month features are negatively supported to the predicted mean consumption. In other words, if we increase the minimum energy consumption it will negatively affect the predicted consumption. However, the outcome of the LR explainer is arguably less informative as the LIMREF output. The DT explainer cannot answer this question. 

\paragraph{What are the hypothetical conditions that would result in obtaining the predicted consumption 568.93kWh? }

Only LIMREF is capable of answering this question. For that, it uses the Hypothetically supporting rules $\Re^+_{h\alpha}$ as follows:
\textit{The conditions that need to be satisfied to maintain the monthly predicted consumption would be mean consumption > 14.74 \& min consumption > 2.88.} 

\textit{Advantage}: By observing these rules, the customers can understand what they need to follow to maintain the current energy consumption if they are happy with their electricity bill. These rules are actionable. 

\paragraph{What are the hypothetical conditions that would result in increasing or decreasing the predicted consumption 568.93kWh? (Counterfactual explanations)}

Both LIMREF and DT explainers are able to answer the above question. 
The LIMREF approach uses the hypothetically contradicting rules ($\Re^-_{h\beta}$) to provide counterfactual explanations: \textit{If you have 6.01 < average temperature $\leq$ 6.31 \& mean consumption > 14.74 \& max consumption > 24.02 it will increase your consumption by 110.64kWh. To reduce your consumption by 216.6kWh the conditions would be 6.01 < average temperature $\leq$ 6.31 \& 15.94 < max consumption $\leq$ 24.02 \& 0.00 $\leq$ min
 consumption $\leq$ 2.88.}

\textit{Advantage}: The customers should try to avoid these practises which may increase the given prediction. In particular, the customer should not exceed the mean consumption greater than 14.74kWh and maximum consumption greater than 24.02kWh when the average temperature is within the given range. 

For the DT, according to Figure~\ref{fig:dt_pic}, if the mean consumption is > 19, there will be an increment of the electricity consumption of the considered household. In contrast, if the mean consumption < 13, the electricity consumption of the selected household will be decreased. 

\subsubsection{Quantitative analysis of the local explainers}

In this section, we analyse fidelity and accuracy of the local explainers. For this analysis, we consider three error measures that are widely used in forecasting and regression.   
Since in the competition, the evaluation process was based on Relative Absolute Error (RAE), in our experiments we consider RAE as an error metric. Further, we consider Root Mean Squared Error (RMSE) and Mean Absolute Error (MAE) error to increase the reliability of our results.

Table~\ref{tab:performance_local} shows the fidelity and the accuracy of the local explainer models by calculating the error between the explainer and the global model (fidelity) and the real values (accuracy), respectively. Here, low error rates denote high fidelity and high accuracy.  We see that the LR model achieves the highest fidelity by obtaining the least error across all three error metrics. LIMREF is second, and DT third in this comparison. 
With regards to the accuracy aspect, our approach outperforms both LR and DT explainers. 

\begin{table}[!htbp]
	\caption{Fidelity and the accuracy measures of the local explainers} 
	\label{tab:performance_local}
	\resizebox{\columnwidth}{!}{%
		\begin{tabular}{c|ccc|ccc}
			\toprule
			{Performance} &  \multicolumn{3}{c}{Fidelity} & \multicolumn{3}{|c}{Accuracy}\\
			\midrule
			Explainer & RAE & RMSE & MAE & RAE & RMSE & MAE \\ \hline
			LR & \textbf{0.348}& \textbf{73.528}	&  \textbf{45.454} & 1.285& 269.277	&  183.263   \\ 
			DT &0.425 &94.506	& 55.491&1.283 &268.233	&182.921\\ 
			LIMREF & 0.390	&  79.522 & 50.945 & \textbf{1.259}	&  \textbf{257.605} & \textbf{179.552}  \\ 
			\bottomrule
		\end{tabular}
	}
\end{table}

In addition to the fidelity and the accuracy aspects, it is interesting to examine the feature importance of the surrogate features across all the instances in the dataset. Therefore, for LR and DT explainers we calculate the feature importance for each instance and average across the instances, see Table~\ref{tab:feature_importance}.

\begin{table}[!tbh]
	\centering
	\caption{Average feature importance of the explainers (LR, DT), and average feature usage frequency for each rule type of LIMREF.} 
	\label{tab:feature_importance}
	\scalebox{0.9}{
		\begin{tabular}{lcc|cccc}
			\hline
			Feature&LR & DT & $\Re_{c\alpha}^+$ & $\Re_{c\beta}^-$ & $\Re^+_{h\alpha}$ & $\Re^-_{h\beta}$\\ 
			\hline
			Max\_cons &1.565 &2.034 & 0.395 & 0.587 & 0.399 & 0.610 \\ 
			Mean\_cons &0.785 &2.299 & 0.357 & 0.526 & 0.368 & 0.611 \\ 
			Min\_cons &7.721 &0.387 & 0.505 & 0.499 & 0.610 & 0.610 \\ 
			Month &2.140 &0.079 & 0.516 & 0.495 & 0.556 & 0.688 \\
			Temp &2.314 &0.643 & 0.372 & 0.624 & 0.606 & 0.658 \\  
			\hline

			
		\end{tabular}
	}
\end{table}

According to Table~\ref{tab:feature_importance}, in the LR explainer, knowing the minimum electricity consumption is considered as the most important feature while the maximum electricity consumption is considered as the least important feature when providing explanations. However, when providing explanations based on the DT explainer knowing the mean consumption is considered as the most important factor whereas the forecasted month has the least importance.  


Table~\ref{tab:feature_importance} furthermore illustrates how often the LIMREF explainer uses the surrogate features when providing explanations for each rule type. It shows that for $\Re_{c\alpha}^+$, and 	$\Re^+_{h\alpha}$ rules, the minimum consumption, month, and the temperature features are used most frequently while rule types $\Re_{c\beta}^-$ and $\Re^-_{h\beta}$ use the features mostly equally. 

\subsection{Model-Agnostic Global Explanations}
We also perform an analysis to provide a global explanation of the forecasting model, to assess whether local explanations are needed in the first place, or a global explanation may be sufficient. Therefore, first, we generate the surrogate feature set from the whole actual dataset using the procedure for surrogate feature generation outlined earlier. After obtaining global explanations we fit LR, DT, and LIMREF explainers on the generated dataset of surrogate features and the global model forecasts. We call these models $\text{LR}_g$, $\text{DT}_g$ and $\text{LIMREF}_g$ in the following. 

%

\begin{table}[!htbp]
	\caption{Fidelity and accuracy of the global explainers} 
	\label{tab:performance_global}
	\resizebox{\columnwidth}{!}{%
		\begin{tabular}{c|ccc|ccc}
			\toprule
			{Performance} &  \multicolumn{3}{c|}{Fidelity} & \multicolumn{3}{c}{Accuracy}\\
			\midrule
			Explainer & RAE & RMSE &  MAE & RAE & RMSE &  MAE \\ 
		\hline
		$\text{LR}_g$ & \textbf{0.355} &\textbf{74.157}	&  \textbf{46.388}  & 1.294 &274.493	&  184.468 \\ 
		 $\text{DT}_g$ & 0.367 & 76.094	& 47.897  & 1.310 & 274.033	& 186.807\\ 
		$\text{LIMREF}_g$	& 0.522 &95.211 & 68.238 & \textbf{1.233} &\textbf{ 257.949} & \textbf{175.862}\\ 
			\bottomrule
		\end{tabular}
	}
\end{table}

According to Table~\ref{tab:performance_global}, $\text{LR}_g$ approximates the global model forecasts best by acquiring the highest fidelity while $\text{LIMREF}_g$ achieves highest accuracy. 
Comparing Tables \ref{tab:performance_local} and \ref{tab:performance_global}, we see that the local explainer models perform better than the global explainers in terms of both fidelity and accuracy. 

\section{Limitations}
Although the proposed LIMREF framework produces model-agnostic rule-based local explanations for the forecasts produced by the global forecasting models, it certainly has limitations. The main limitation of the current work is that the chosen application is limited by the competition dataset that we considered in our experiments. As such, we could only use the available features, and, more importantly, the forecasting horizon was very long (a full year). As such, the surrogate model's features are limited through the dataset and problem statement and can only consider past consumption and average temperatures on a relatively high aggregation level. 
Also, the four types of explanations are a direct mapping from the theory, of what the algorithm
generates. As they are quite general, they may not be specific enough for a particular application.

\section{Conclusions and Future Work}
Electricity demand forecasting is important for many applications, and global forecasting models that train across series have been proposed to perform this task accurately. However, accurate forecasts are oftentimes not enough to make informed and good decisions. For example, customers do not know what they should and should not do to decrease their electricity consumption.
In this paper, we propose the LIMREF framework, which is a method developed for the  
FUZZ-IEEE competition on Explainable Energy Prediction
and also, to the best of our knowledge, the first model-agnostic rule-based algorithm which explains global forecasting models locally. LIMREF generates six types of guidances in the form of rule-based explanations for global forecasting models. These rule-based guidances could be communicated to energy customers, e.g., in their invoices, to help them understand their bills and what they can potentially do to lower their energy costs. We have compared LIMREF with two other approaches regarding fidelity, accuracy, and comprehensibility, showing its unique strengths and competitiveness.

In the future, we hope to validate the robustness of the LIMREF framework by applying it to different forecasting applications outside of the particular energy forecasting
application case. Moreover, as with other ML models, this model is susceptible to biases in the data. However, it is commonly hypothesized that post-hoc explanations can assist in detecting model biases. Therefore, our explainer model can be able to help uncover such biases. This has not been addressed in the current paper but would be an excellent research question for future work. Furthermore, the ``right'' level of locality for explanations is largely unexplored in the time series forecasting domain, which can be explored more in the future.

%
\section{Acknowledgements}
We sincerely thank Hansika Hewamalage for supporting us in the global modelling part by replicating the global forecasting model developed for the previous competition,
IEEE-CIS technical challenge on Energy Prediction from Smart Meter Data on the same dataset. 
We would also like to thank Isaac Triguero and his team for organising the 
FUZZ-IEEE competition on Explainable Energy Prediction. This research was supported by the Australian Research Council under grant DE190100045.

\newpage
\bibliography{dilini2021limref}

\end{document}